\documentclass[conference]{IEEEtran}
\IEEEoverridecommandlockouts
\usepackage{amsmath,amssymb,amsfonts}
\usepackage{algorithmic}
\usepackage{graphicx}
\usepackage{textcomp}
\usepackage[table]{xcolor}
\usepackage{multirow}
\usepackage{url}
\usepackage{hyperref}
\usepackage[capitalize]{cleveref} 
\usepackage{tikz}
\usepackage{tabularx}
\usepackage{svg}
\usepackage{enumitem}

\usetikzlibrary{arrows.meta,positioning,fit,calc}

\newcommand{\BT}{\mathcal{T}}
\newcommand{\NLinstruction}{\mathcal{L}}
\newcommand{\SkillLibrary}{\mathcal{K}}
\newcommand{\skill}{k}
\newcommand{\Rootstocks}{\mathcal{R}}
\newcommand{\rootstock}{r}
\newcommand{\BToperators}{\mathcal{O}}
\newcommand{\btoperator}{o}
\newcommand{\Contract}{\mathcal{C}}

\newcommand{\scoregreen}[1]{\textcolor{green!60!black}{#1}}
\newcommand{\scoreorange}[1]{\textcolor{orange!80!black}{#1}}
\newcommand{\scorered}[1]{\textcolor{red!70!black}{#1}}

\usepackage[style=ieee,maxnames=3,minnames=1]{biblatex}
\begin{document}
\addtolength{\topmargin}{6pt}

\title{
Contract-Grounded Behavior Tree Synthesis \\ via Coding Agent
}

\author{
Jonathan Salfity$^{1}$, Robert Blake Anderson$^{1}$, and Mitch Pryor$^{1}$
\thanks{$^{1}$Nuclear and Applied Robotics Group, Department of Mechanical Engineering ,
        The University of Texas at Austin,
The project website can be found at: \url{https://jsalfity.github.io/agentic-bt-gen-webviewer/}
}
}

\maketitle

\begin{abstract}
Synthesizing deployable robot behavior trees (BTs) from natural language (NL) requires \emph{grounding} to ensure every generated BT references only skills a robot can actually execute.
Existing LLM-based BT synthesis approaches often place this grounding responsibility on the prompt author. 
This makes deployment brittle when the author does not know which skills the robot can execute, how those skills are parameterized, or how the robot runtime software constrains valid BT structure.
This paper proposes a contract-grounded BT synthesis architecture in which a coding agent queries a robot-side Model Context Protocol (MCP) server to retrieve an explicit contract consisting of a skill library, permitted BT operators, and optional BT composition templates, before synthesizing a BT for validation and execution.
In our framework, non-expert operators issue NL commands without knowledge of robot implementation details, while a robot runtime validation gate enforces correctness before execution.
We evaluate two LLMs, a closed model (Sonnet 4.6) and a smaller open-source model (Gemma4:31b), across 110 simulated tasks in PyRoboSim and 14 tasks on a physical Husarion Panther robot.
Results show that contract grounding enables near-perfect BT validation and high task success, that BT composition templates substantially recover success on reactive control-flow tasks for the smaller model, and that the architecture transfers to physical hardware running a Nav2 stack opaque to both operator and agent.
\end{abstract}


\section{Introduction}\label{sec:introduction}
Robots are increasingly expected to execute tasks specified by non-expert users using high-level natural language (NL) commands.
Robots can greatly benefit from interpretable, safety-aware control policies with explicit action sequencing and failure handling that are difficult to guarantee from NL alone.
However, directly translating task command intent into reliable, explicit robot behavior remains difficult. 
This gap motivates representations that can accept flexible NL inputs while producing behavior control logic that is explicit, inspectable, and constrained to the robot's capabilities.

Behavior Trees (BTs) are a widely used representation for reactive, modular robot control~\cite{BT_in_robotics_AI_Colledanchise2018,BT_in_RobotControlSystems_Ogren_Sprague_2022,iovino2020surveybehaviortreesrobotics}.
Unlike opaque neural network end-to-end policies, BTs \emph{explicitly} encode control flow and recovery logic, enabling operators and developers to inspect and verify high-level behavior before, during, and after execution.
Despite these advantages, manually constructing BTs remains labor-intensive as authoring requires knowledge of the robot's available skills, their parameter interfaces, and how to compose them into correct reactive structures.
This is an expertise that non-expert operators typically do not have~\cite{iovino2020surveybehaviortreesrobotics}.

Recent efforts have explored using Large Language Models (LLMs) to synthesize BTs directly from natural language through prompt-based methods, fine-tuned models, hybrid symbolic planners, and agentic pipelines~\cite{LLM_as_BT_planner_2025,lykov2023llmbrainaidrivenfastgeneration,LLM_OBTEA_ijcai2024,BETR_XP_LLM_Styrud_2025}.
These approaches reduce manual authoring efforts, but cannot ensure the synthesized BT references skills the robot can actually execute. 
This check is typically the prompt author's responsibility, as robot runtime software in existing LLM-based BT methods generally do not impose strict rules on what an LLM may generate.
For precompiled, opaque, or evolving robot stacks, prompt-authoring becomes brittle. 
Prior work shows that LLM-generated robot programs can fail through invalid arguments, infeasible actions, and misuse of robot function semantics~\cite{CodeBotler}.

Coding agents such as Anthropic Claude Code~\cite{anthropic_claude_code} and OpenAI Codex~\cite{chen2021codex} offer an emerging paradigm of code synthesis.
Unlike a standalone language model that generates output from a fixed prompt, coding agents decompose tasks, interactively query tools to retrieve context, and are generally tasked to synthesize structured code output from their natural language input.

This paper proposes a \emph{contract-grounded} coding agent architecture for BT synthesis, where \emph{grounding} requires every synthesized BT to reference only skills the robot can execute with valid parameters and allowable control structures.
The agent queries a robot-side Model Context Protocol (MCP) server~\cite{anthropic_mcp_2024} to retrieve an explicit contract before synthesizing a BT for validation and execution.
Specifically, this effort provides a framework to enable:
\begin{enumerate}[nosep]
    \item \textbf{Contract-grounded BT synthesis:} A coding agent synthesizes BTs from NL under an explicit contract that specifies skills, typed parameters, permitted BT operators, and optional BT composition templates;
    \item \textbf{Robot-runtime authority:} Contract information exposed through a server-mediated interface, while implementation code, BT instantiation, validation, and execution authority remain inside the robot runtime (the on-robot BT execution software stack).
    \item \textbf{Quantitative evaluation in simulation:} We evaluate episodic BT synthesis quality in PyRoboSim~\cite{pyrobosim_docs} using pre- and post-execution metrics for structured and language-varied tasks; and
    \item \textbf{Physical deployment:} We deploy the architecture on a physical robot with precompiled, opaque navigation and perception modules, showing a contract-grounded interface applies across distinct BT runtimes and skills.
\end{enumerate}


\section{Related Work}\label{sec:related_work}

BTs are a modular, hierarchical, and reactive control architecture that are adopted widely in robotics~\cite{BT_in_robotics_AI_Colledanchise2018,iovino2020surveybehaviortreesrobotics}.
BTs are well suited to robotics applications where operators or developers benefit from explicit, inspectable knowledge of a robot's high-level control logic before and during execution.
Despite their advantages, BTs are still commonly authored manually, and this process is difficult \cite{iovino2020surveybehaviortreesrobotics}.
The designer must know which skills are available, explicitly encode task-switching logic across BT nodes, anticipate how failure statuses propagate up the tree, and select correct parameters for each skill -- a process that requires simultaneous knowledge of the robot's capabilities and BT design conventions~\cite{iovino2020surveybehaviortreesrobotics,BT_in_robotics_AI_Colledanchise2018}. 

Prior efforts to automate BT synthesis include symbolic planning and formal verification~\cite{synthesis_LTL_Colledanchise_Murray_Ogren_2017,blended_reactive_planning_Colledanchise2019,eBT_Rovida_Grossmann_Kruger_2017,Paxton2019RepresentingRT}; learning-based algorithms including reinforcement learning, evolutionary methods; and learning from demonstration~\cite{banerjee_autonomous_2018,colledanchise_learning_2019,french_learning_2019}. 
Refer to~\cite[Sec.~4]{iovino2020surveybehaviortreesrobotics} for a comprehensive pre-LLM survey.
Formal methods offer correctness guarantees but require expert domain modeling in formal languages and cannot interpret high-level natural language.
Learning-based approaches require significant domain-specific training data, making rapid one-shot synthesis from a novel task description impractical.
By contrast, our method utilizes a general coding agent that is pre-trained to interpret NL and, as we show, is able to adapt to the BT construction domain.

More broadly, LLM-based robot program generation has been grounded through prompted APIs, candidate skill sets, and execution-time feedback~\cite{codeaspolicies2022,ProgPrompt,SayCan,CodeBotler}.
These approaches typically validate individual actions or post-execution outcomes, rather than checking the composed control-flow structure itself against a permitted operator set before execution.

LLM-integrated BT synthesis methods can be organized into five categories.
\textit{Prompt-based} methods~\cite{LLM_as_BT_planner_2025,BT_generation_cao,BETR_XP_LLM_Styrud_2025,zhang_codebt_2025} use carefully crafted prompts and in-context examples to induce BT structure from the LLM; correctness depends heavily on prompt design and the model's implicit knowledge of available skills.
Code-BT~\cite{zhang_codebt_2025} improves pass rate by routing generation through rule-constrained Python code parsed via an Abstract Syntax Tree (AST) into XML BTs, but still relies on a designer-maintained API library supplied in the prompt and multi-round iterative feedback rather than one-shot synthesis.
\textit{Dialogue-driven} methods~\cite{DiaGBT_2024,llmmarslargelanguagemodel} iterate on BT synthesis through multi-turn interaction, gathering context from the user across turns; grounding depends on what the user supplies rather than a robot-side interface.
\textit{Fine-tuned} models~\cite{lykov2023llmbrainaidrivenfastgeneration,izzo2024btgenbotbehaviortreegeneration} train on curated BT datasets to improve output consistency, but require retraining when robot capabilities or operator sets change.
\textit{Hybrid symbolic-LLM} frameworks~\cite{LLM_OBTEA_ijcai2024} combine LLMs with symbolic planners to improve reliability, but still require domain modeling and formal goal specification.
\textit{Agentic} methods~\cite{li2024studytrainingdevelopinglarge} decompose BT generation into multiple modules, such as goal interpretation, planning, verification, and structured BT construction; however, they do not actually deploy their method on a robot, either simulated or physical.

\textbf{Positioning of this work:}
Prior LLM robot-programming and BT synthesis systems place grounding responsibility on a designer-maintained generation interface, training pipeline, symbolic model, user dialogue, or agent-side module; however, this dependency breaks when the robot stack is precompiled, opaque, or unavailable to the prompt author at deployment time.
This work moves grounding into a server-mediated robot interface where an agent queries the robot-side contract before synthesis, and the runtime validates the generated BT before execution, enabling non-expert operators to issue NL commands without implementation details, while the validation gate enforces correctness.

\section{Problem Definition}\label{sec:main_problem}
We consider the problem of synthesizing deployable robot BTs from natural-language commands in a setting where the operator knows task intent but not the robot's underlying implementation details.
We formulate the task as mapping a natural-language instruction, $\NLinstruction$, into a structured BT, $\BT$, through a synthesis method, $\mathcal{S}$:
\begin{equation}
    \label{eq:problem}
    \BT = \mathcal{S}(\NLinstruction, \Contract)
\end{equation}
where $\Contract = (\SkillLibrary, \BToperators, \Rootstocks)$ is the robot-side contract, a machine-readable interface exposed by the robot that provides information for high-level BT composition without requiring access to backend implementation code.
The contract consists of:
\begin{enumerate}[nosep]
    \item \textbf{Skill Library, $\SkillLibrary$:} the names, parameter schemas, and intended semantics of executable robot skills, $\skill \in \SkillLibrary$.
    \item \textbf{Allowed BT Operators, $\BToperators$:} the exclusive set of BT control-flow operators, $\btoperator \in \BToperators$, that are allowed to appear in the synthesized tree, and
    \item \textbf{BT Rootstocks, $\Rootstocks$:} we introduce the term \emph{rootstock}, $\rootstock \in \Rootstocks$, to refer to templated BT fragments that encode BT composition patterns for recurring task structures.
    For example, a \texttt{search\_pick\_selector} rootstock encodes a selector-branch that tries each candidate location in sequence, detects and picks an object if found, and falls back to the next location otherwise.
\end{enumerate}

The synthesized BT must satisfy two core requirements before executing on the robot:
\begin{enumerate}[nosep]
    \item \textbf{Grounded composition:} every leaf node must correspond to an exposed, executable skill, $\skill \in \SkillLibrary$, with valid parameters under the skill schema; and
    \item \textbf{Constrained control structure:} the synthesized tree must be composed only from the BT operators, $\btoperator \in \BToperators$, explicitly permitted by the contract.
\end{enumerate}

Unlike prompt-based methods that require the author to maintain grounding information upfront~\cite{LLM_as_BT_planner_2025,zhang_codebt_2025}, moving grounding into a server-mediated robot interface allows a non-expert operator to express \emph{what} the robot should do while the contract supplies the \emph{how} for synthesis.

\section{Approach}\label{sec:approach}
Our approach realizes the robot-side contract formulation in \Cref{sec:main_problem} with a client-server architecture for constrained BT synthesis.
Inspired by \cite{ros_mcp_server}, we utilize a robot-side MCP server that exposes a machine-readable contract to a coding agent, where the robot server is the authority of exposed information for synthesis and validation prior to BT execution.

\begin{figure}[h!]
    \centering
    \includegraphics[width=\columnwidth]{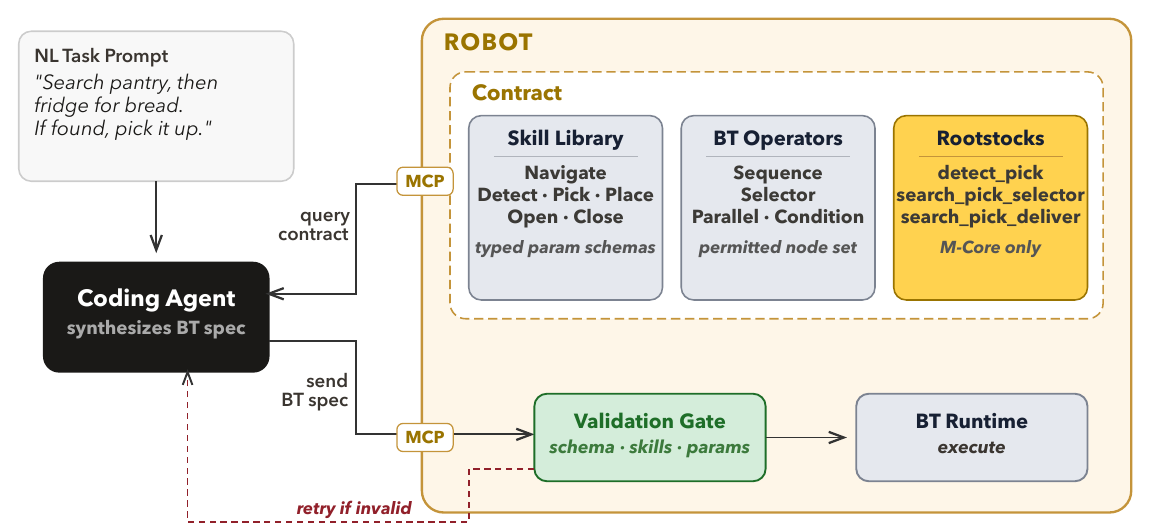}
    \caption{\textbf{System overview:} A coding agent queries the robot-side contract, $\Contract=(
        \SkillLibrary, \BToperators, \Rootstocks)$ through MCP, synthesizes a BT, and submits it for validation and execution.}
    \label{fig:approach}
\end{figure}

\subsection{System Overview}
\label{subsec:approach_overview}
\Cref{fig:approach} summarizes the overall client-server architecture and single-task synthesis pipeline.
Given a natural-language instruction, $\NLinstruction$, the agent first queries the robot-side MCP server to retrieve the exposed contract, $\Contract$.
Using only the contract, the agent synthesizes a BT, $\BT$, and submits it to the robot via an MCP tool first for validation and then execution.
If the BT fails validation, the agent receives structured error feedback and retries; once valid, the robot runtime executes the BT.

This paper studies episodic, single task synthesis: the agent produces a single BT per instruction without online replanning, isolating synthesis interface quality for the quantitative evaluation in \Cref{sec:experiments}.

\subsection{Contract Interface \& Contents}\label{subsec:contract_interface}
The architecture separates the public synthesis interface from the private execution backend.
In practice, a robot-side developer authors the contract before deployment, including any rootstocks exposed to the agent.
The MCP server exposes five tools: \texttt{get\_skill\_library}, \texttt{get\_bt\_operators}, \texttt{get\_rootstocks}, \texttt{get\_world\_vocabulary} and \texttt{send\_to\_robot}, through which the agent queries the contract and submits BTs for validation and execution.
This interface constrains the agent to behavior policy synthesis, allowing it to select skills, fill parameters, and arrange BT control flow, but not modify skill implementations or bypass the robot runtime.
\Cref{tab:contract_resources} shows exposed elements in a contract for the simulated robot used in \Cref{sec:experiments}.
Additionally, each skill exposes typed parameter schemas as well as success, running, and failure conditions.
\newcommand{\kv}[2]{\textbf{#1}~#2}%
\begin{table}[h!]
\caption{PyRoboSim Contract}
\label{tab:contract_resources}
\centering
\scriptsize
\setlength{\tabcolsep}{2pt}
\renewcommand{\arraystretch}{1.06}
\begin{tabular}{p{0.32\columnwidth}p{0.32\columnwidth}p{0.32\columnwidth}}
\hline
\textbf{Skill Library, $\SkillLibrary$}
& \textbf{BT Operators, $\BToperators$}
& \textbf{Rootstocks, $\Rootstocks$} \\
\hline
\begin{minipage}[t]{0.32\columnwidth}
\ttfamily\scriptsize
\textbf{navigate}:\\
\hspace*{0.5em}\kv{desc:}{path to target loc/pose}\\
\hspace*{0.5em}\kv{params:}{target\_loc, pose, path}\\
\hspace*{0.5em}\kv{outputs:}{result}\\[3pt]
\textbf{detect}:\\
\hspace*{0.5em}\kv{desc:}{...}\\
\hspace*{0.5em}\kv{params:}{...}\\
\hspace*{0.5em}\kv{outputs:}{...}\\[3pt]
\textbf{pick}:\\ \hspace*{0.5em}...\\
\textbf{place}:\\ \hspace*{0.5em}...\\
\textbf{open}:\\ \hspace*{0.5em}...\\
\textbf{close}:\\ \hspace*{0.5em}...
\end{minipage}
&
\begin{minipage}[t]{0.32\columnwidth}
\ttfamily\scriptsize
\textbf{sequence}:\\
\hspace*{0.5em}\kv{memory:}{bool}\\[3pt]
\textbf{selector/fallback}:\\
\hspace*{0.5em}\kv{memory:}{bool}\\[3pt]
\textbf{action}:\\
\hspace*{0.5em}\kv{action:}{skill\_name}\\[3pt]
\textbf{condition}:\\
\hspace*{0.5em}\kv{key:}{bb\_key}\\
\hspace*{0.5em}\kv{op:}{==, !=, in, ...}\\
\hspace*{0.5em}\kv{value:}{any}\\[3pt]
\textbf{parallel}:\\ \hspace*{0.5em}...\\
\textbf{decorator}:\\ \hspace*{0.5em}...
\end{minipage}
&
\begin{minipage}[t]{0.32\columnwidth}
\ttfamily\scriptsize
\textbf{detect\_pick}:\\
\hspace*{0.5em}\kv{desc:}{detect-gated pick}\\
\hspace*{0.5em}\kv{slots:}{obj\_query, src\_loc}\\[3pt]
\textbf{search\_pick\_}\\
\textbf{selector}:\\
\hspace*{0.5em}\kv{desc:}{...}\\
\hspace*{0.5em}\kv{slots:}{...}\\[3pt]
\textbf{search\_pick\_}\\
\textbf{deliver}:\\
\hspace*{0.5em}\kv{desc:}{...}\\
\hspace*{0.5em}\kv{slots:}{...}
\end{minipage}
\\\\
\hline
\end{tabular}
\end{table}

\subsection{Validation Gate}\label{subsec:validation}
Before executing the BT on the robot, the synthesized BT is checked against the contract and tested for BT syntax.
Validation verifies that the specification parses as a BT, uses only permitted operators, grounds all leaf nodes as skills in the schema, and supplies type-compatible parameters.
As shown in \Cref{fig:approach}, invalid BTs are sent back to the agent with specific BT validation errors (BT syntax, hallucinated skills, unallowed parameterization, etc.), where the agent can \textit{retry} BT synthesis and submission to the robot.
Validation ensures that only BTs that pass these checks are instantiated by the backend runtime.
\section{Experiments}
\label{sec:experiments}
Our experiments evaluate whether the contract-grounded synthesis approach proposed in \Cref{sec:approach} produces valid and executable robot BTs from natural language, and whether synthesized BTs can accomplish a commanded task when deployed.
Our evaluation asks and answers three questions:
\begin{enumerate}[label=\textbf{Q\arabic*:}, ref=Q\arabic*, nosep]
    \item \label{q:contract_grounding} \textbf{Contract grounding} Does the proposed skill contract expose the grounding information needed for valid and executable BT synthesis?
    \item \label{q:composition_priors} \textbf{Composition priors} How do rootstock templates affect BT synthesis across task archetypes and language variations?
    \item \label{q:physical_deployment} \textbf{Physical deployment} Does the contract support deployable BT synthesis on a physical robot running an opaque Nav2 stack with complex ROS2 data interactions?
\end{enumerate}

We evaluate the first two questions broadly and quantitatively in PyRoboSim and the third question on hardware.
Following benchmark-oriented evaluations of NL for program synthesis, such as in \cite{CodeBotler,chen2021codex}, we adopt first-attempt validation metrics, execution-based success checks, and perform a failure analysis.

\subsection{Evaluation Domains}\label{subsec:evaluation_domains}
\textbf{PyRoboSim} \cite{pyrobosim_docs} is the primary benchmark domain, with PyTrees \cite{pytrees} used as the BT execution framework with a custom JSON wrapper for BT specification.
There was no high-level BT specification for PyTrees at the time of writing, so we wrote our own.
The exposed skill set consists of PyRoboSim's native high-level mobile-manipulation actions \texttt{Navigate}, \texttt{Detect}, \texttt{Pick}, \texttt{Place}, \texttt{Open}, and \texttt{Close}, etc. shown briefly in \Cref{tab:contract_resources}.
In our experiments, the client receives the full skill schema for six deterministic actions, the allowable PyTrees BT node and control-flow set, and a live world vocabulary containing named objects, object categories, rooms, locations, object spawns, and hallways — but not ground-truth relationships.

We evaluated three suites in PyRoboSim:
The \textbf{Core60} suite tests structured BT composition over task archetypes such as sequential manipulation, selector-style search, search-and-place, pick-and-return, and pick-and-place.
The \textbf{Lang50} suite tests robustness to natural-language variation by preserving task semantics with varied paraphrases, lexical choices, clause ordering, and constraint phrasing.
The Lang50 suite is inspired by recent work \cite{WannaLetsTA} showing that recent learning-based end-to-end policies are brittle to language-varied task commands. 
Representative task archetypes and language-variation categories are shown in \Cref{tab:pyrobosim_core60_results} and \Cref{tab:pyrobosim_lang50_results}.

The \textbf{OOC10} suite tests out-of-contract behavior with ten task prompts that require capabilities absent from the exposed skill set such as speech, unallowable environment actuation, continuous following, or object counting.
OOC10 samples from the effectively unbounded space of out-of-contract requests to probe whether the agent recognizes the contract boundary.
The same ten prompts are evaluated in both the PyRoboSim and physical robot domains; success is defined as the system \emph{not} attempting to produce a BT.
(Refer to \Cref{subsec:metrics} for validation and success metric definitions.)
OOC10 results across both domains are reported in \Cref{sec:experiments:ooc_results}.

\textbf{Physical Robot Study} tests whether our method supports deployable BT synthesis when the underlying robot infrastructure is largely precompiled and opaque to both the operator and the agent.
We instantiate the architecture on a Husarion Panther mobile robot running ROS2 \cite{ROSQuigley2009} with a precompiled Nav2 \cite{nav22020} navigation stack, a YOLO-based \cite{ultralytics_yolo} detection node, and a custom heading controller.
The BTs are orchestrated by BT.CPP \cite{btcpp} and defined via the native BT.CPP BT XML specification.
This robot stack is structurally distinct from the PyRoboSim simulation domain in both BT runtime (BT.CPP vs. PyTrees), skill semantics, and BT specification format (XML vs. JSON).

The Panther skill library exposes hooks into opaque, pre-behaviorized Nav2 actions (\texttt{Navigate}, \texttt{Spin}, \texttt{Wait}, \texttt{MoveForward}, \texttt{BackUp}) and custom perception skills (\texttt{WaitForDetection}, \texttt{CheckDetection}, \texttt{HeadingController}).
The allowable BT composition operators are the common BT.CPP control flow operators.
Rootstock templates such as \texttt{patrol\_and\_return} and \texttt{spin\_until\_detect} encode reactive fallback patterns.


\subsection{Methods and Models Compared}\label{subsec:methods_compared}
\textbf{Methods:} For both studies, we compare two primary interface configurations and two models, summarized in \Cref{tab:method_definitions_restructured}.
\textbf{M-Core} is the full method from \Cref{sec:approach}: the coding agent queries the full robot-side contract $(\SkillLibrary, \BToperators, \Rootstocks)$ before synthesizing a BT specification for validation and execution.
\textbf{B1} uses the same contract-grounded interface but exposes only $(\SkillLibrary, \BToperators)$, isolating the contribution of reusable BT composition priors.

\begin{table}[h]
\caption{Overall experimental design across suites, methods, and models. \checkmark~= evaluated; \texttimes~= not evaluated.}
\label{tab:method_definitions_restructured}
\centering
\scriptsize
\setlength{\tabcolsep}{4pt}
\renewcommand{\arraystretch}{1.2}
\begin{tabular}{lcccc}
\hline
& \multicolumn{2}{c}{\textbf{M-Core}} & \multicolumn{2}{c}{\textbf{B1}} \\
\cline{2-5}
\textbf{Suite} & \textbf{Sonnet 4.6} & \textbf{Gemma4:31b} & \textbf{Sonnet 4.6} & \textbf{Gemma4:31b} \\
\hline
PyRoboSim Core60  & \checkmark & \checkmark & \checkmark & \checkmark \\
PyRoboSim Lang50  & \checkmark & \checkmark & \checkmark & \checkmark \\
PyRoboSim OOC10   & \checkmark & \checkmark & \texttimes  & \texttimes  \\
Panther 14     & \checkmark & \checkmark & \checkmark & \checkmark \\
Panther OOC10    & \checkmark & \checkmark & \texttimes  & \texttimes  \\
\hline
\end{tabular}
\end{table}

We do not treat prompt engineering as a primary experimental method because it assumes that the prompt author already knows the skill schema, BT schema, symbolic vocabulary, and useful composition patterns.
We therefore note prompt-only generation is a \emph{full-information} reference, a scenario we are not targeting, rather than a main baseline for the deployment setting targeted in this paper.

\textbf{System Level Prompt \& Models:}
Each BT is generated in an isolated Claude Code subprocess via \texttt{claude -p} with no shared context between tasks.
The model is instructed via a system prompt: \emph{Query the robot MCP server for the skill library, BT schema, and rootstocks, (if available), then construct a valid BT and send it to the robot. The robot task is to $<$insert task prompt$>$}.

Both sets of experiments, simulated and physical, compare generated BTs using two LLMs: \textbf{Sonnet 4.6} \cite{anthropic_claude_sonnet}, a closed, paid model, and \textbf{Gemma4:31b} \cite{gemma4}, a much smaller, open-source model served locally via Ollama \cite{ollama}.
Each LLM serves as the reasoning model wrapped in the terminal-based Claude Code application. 
These two models represent opposite ends of the current deployment spectrum. 
Sonnet 4.6 is a large, closed, commercially-hosted model, while Gemma4:31b is a significantly smaller open-source model that can be served locally on a single workstation. 
We use both to assess whether the contract-grounded interface is robust across this range.

\subsection{Metrics}\label{subsec:metrics}
\textbf{Valid@1}, a pre-execution metric, measures whether an agent produces a syntactically correct BT on the agent's first BT submission, e.g. the first \texttt{send\_to\_robot} tool call.
Valid@1 is a binary metric: 1 if the agent's first BT submission passes validation, 0 otherwise --- analogous to \emph{pass@1} in code-generation benchmarks~\cite{chen2021codex} but applied as a deterministic per-task check rather than a probability estimate over samples.
We define a BT as \textit{valid} if it is well-formed XML (BT.CPP v4) or JSON (custom PyTrees specification), references only skills from the robot's skill library, and supplies all required skill parameters drawn from the exposed world vocabulary.
These symbolic names are resolved to world coordinates (or domain-specific identifiers such as Nav2 pose goals and YOLO classes) by the underlying skill implementations at execution time.
In our experiments, we arbitrarily allowed the agent up to 5 retries to produce a valid BT.
Valid@1 captures first-attempt BT validation, while Success (defined below) is computed on the first BT that passes validation within those retries; if a retry is needed, Success is judged on that first valid BT.

\textbf{Success}, a post-execution metric, measures whether a robot accomplished a given task after passing validation.
In simulation, success is strictly determined by comparing the world state before and after execution against a predefined, task-specific success condition. 
As an example, for a commanded task, \textit{``Search pantry, then desk for snacks. If found, pick it up and return to dining,''}, in a PyRoboSim world where snacks can be found, success requires the robot to be holding snacks and located in the dining room. 
In a PyRoboSim world where snacks cannot be found, success requires the robot to end in the dining room.
On physical hardware, success is assessed by human observation of whether the robot achieved its commanded goal.


Our \textbf{Failure Taxonomy} breaks failures into three categories:
\begin{itemize}[nosep]
    \item \emph{Never Valid BT}, a pre-execution failure, captures tasks for which the agent never produced a BT that passes the static contract check within our arbitrary 5-retry budget, implying no BT is executed. 
    \item \emph{Runtime Failure} occurs when a BT executes but reaches PyRoboSim's \textsc{Failure}, \textsc{Timeout}, or \textsc{Error} status, usually due to a logically incorrect BT, such as a robot trying to grasp an object when it is already holding an object.
    \item \emph{Goal Mismatch}, a post-execution failure, occurs when a BT passes static validation and completes execution, but the robot does not achieve the task goal. 
\end{itemize}

\subsection{PyRoboSim Results}\label{sec:experiments:pyrobosim_results}
\Cref{tab:pyrobosim_main_comparison} shows the full ablation study results, where \Cref{tab:pyrobosim_core60_results} and \Cref{tab:pyrobosim_lang50_results} provide a deep dive into the Core60 and Lang50 task suites, respectively. \Cref{tab:pyrobosim_failure_breakdown} provides a three-category failure breakdown.

\begin{table}[h]
\caption{Primary PyRoboSim comparison.}
\label{tab:pyrobosim_main_comparison}
\centering
\small
\renewcommand{\arraystretch}{1.1}
\setlength{\tabcolsep}{4pt}
\begin{tabular}{llcccc}
\hline
\textbf{Model} & \textbf{Method} & \textbf{Suite} & \textbf{N} & \textbf{Valid@1} & \textbf{Success} \\
\hline
Sonnet 4.6  & B1     & Core60 & 60 & 59 \scoregreen{(98\%)}  & 56 \scoregreen{(93\%)} \\
Sonnet 4.6  & B1     & Lang50 & 50 & 49 \scoregreen{(98\%)}  & 45 \scoregreen{(90\%)} \\
Sonnet 4.6  & M-Core & Core60 & 60 & 60 \scoregreen{(100\%)} & 58 \scoregreen{(97\%)} \\
Sonnet 4.6  & M-Core & Lang50 & 50 & 50 \scoregreen{(100\%)} & 48 \scoregreen{(96\%)} \\
\hline
Gemma4:31b & B1     & Core60 & 60 & 51 \scoreorange{(85\%)}  & 30 \scorered{(50\%)} \\
Gemma4:31b & B1     & Lang50 & 50 & 46 \scoregreen{(92\%)}  & 25 \scorered{(50\%)} \\
Gemma4:31b & M-Core & Core60 & 60 & 54 \scoreorange{(90\%)}  & 50 \scoreorange{(83\%)} \\
Gemma4:31b & M-Core & Lang50 & 50 & 45 \scoreorange{(90\%)}  & 41 \scoreorange{(82\%)} \\
\hline
\end{tabular}
\end{table}

\textbf{Key Insights:} 
For \ref{q:contract_grounding}, Sonnet achieves near-perfect Valid@1 across all conditions, indicating the contract provides enough information to produce well-formed, grounded BTs in our evaluated domains.
Gemma's Valid@1 is lower but the more dramatic gap is in success: Gemma B1 achieves only 50\% on Core60 and 50\% on Lang50, while M-Core recovers these to 83\% and 82\% respectively. 

The task archetype breakdown from \Cref{tab:pyrobosim_core60_results} and language variation in \Cref{tab:pyrobosim_lang50_results} reveal where this gap originates.
Sequential pick is largely solved (92--100\% across all conditions), while search archetypes (selector search, search-and-place, and pick-and-return) concentrate failures.
Gemma B1 collapses on these archetypes (0-42\%), as they require reactive fallback control flow that the model cannot reliably construct from the skill contract alone.
M-Core substantially closes this gap, with rootstock templates providing the correct selector-branch structure, though search-and-place remains the hardest archetype for Gemma M-Core (58\%).

For \ref{q:composition_priors}, we observe rootstocks also benefit both models on structurally demanding variations.
Sonnet B1's constraint wording and clause reordering drop to 80\% and 90\% but recover to 100\% under M-Core; Gemma B1's drop to 20\% and 50\% and partially recover under M-Core to 60\% and 80\%.

\begin{table}[h]
\caption{\textbf{Core60} PyRoboSim Success by task archetype}
\label{tab:pyrobosim_core60_results}
\centering
\footnotesize
\renewcommand{\arraystretch}{1.15}
\setlength{\tabcolsep}{3pt}
\begin{tabular}{>{\raggedright\arraybackslash}p{0.17\columnwidth}>{\raggedright\arraybackslash}p{0.30\columnwidth}cccc}
\hline
 &  & \multicolumn{2}{c}{\textbf{Sonnet 4.6}} & \multicolumn{2}{c}{\textbf{Gemma4:31b}} \\
\cline{3-4}\cline{5-6}
\textbf{Archetype} (n=12) & \textbf{Example prompt} & \textbf{B1} & \textbf{M-Core} & \textbf{B1} & \textbf{M-Core} \\
\hline
Sequential pick  & ``\textit{Go to pantry, detect bread, and pick it up.}''                                        & 100\% & 100\% &  92\% & 100\% \\
Selector search  & ``\textit{Search pantry, then fridge for bread. If found, pick it up.}''                       &  83\% & 100\% &  33\% &  83\% \\
Search-and-place & ``\textit{Search pantry, then fridge for bread. If found, place it on table\_tabletop.}''     &  83\% & 100\% &   0\% &  58\% \\
Pick-and-return  & ``\textit{Search pantry, then fridge for bread. If found, pick it up and return to dining.}'' & 100\% & 100\% &  42\% &  83\% \\
Pick-and-place   & ``\textit{Pick bread from pantry and place it on table\_tabletop.}''                           & 100\% &  83\% &  83\% &  92\% \\
\hline
\end{tabular}
\end{table}

\begin{table}[h]
\caption{\textbf{Lang50} PyRoboSim Success by variation category}
\label{tab:pyrobosim_lang50_results}
\centering
\footnotesize
\renewcommand{\arraystretch}{1.15}
\setlength{\tabcolsep}{3pt}
\begin{tabular}{>{\raggedright\arraybackslash}p{0.17\columnwidth}>{\raggedright\arraybackslash}p{0.30\columnwidth}cccc}
\hline
 &  & \multicolumn{2}{c}{\textbf{Sonnet 4.6}} & \multicolumn{2}{c}{\textbf{Gemma4:31b}} \\
\cline{3-4}\cline{5-6}
\textbf{Variation type} & \textbf{Example prompt} & \textbf{B1} & \textbf{M-Core} & \textbf{B1} & \textbf{M-Core} \\
\hline
Paraphrase (n=10)& ``\textit{Head to pantry, detect bread, and pick it up.}''                                                   & 100\% & 100\% & 100\% & 100\% \\
Lexical substitution (n=20)& ``\textit{Pick bread from pantry and set it on table\_tabletop.}''                                        &  90\% &  90\% &  40\% &  85\% \\
Constraint wording (n=10)& ``\textit{Search pantry, and only if needed then fridge for bread. If found, place it on table\_tabletop.}'' &  80\% & 100\% &  20\% &  60\% \\
Clause reordering (n=10)& ``\textit{Return to dining while holding bread, but only after you have found and picked it up.}''           &  90\% & 100\% &  50\% &  80\% \\
\hline
\end{tabular}
\end{table}

Shown in \Cref{tab:pyrobosim_failure_breakdown}, across both suites, Gemma B1's failures concentrate in Runtime Failures, implying the synthesized BTs raised errors and failures in PyRoboSim.
Gemma B1 also shows the highest Goal Mismatch rate, where the BT runs to completion but does not achieve the goal specification.
Importantly, M-Core reduces both failures, but interestingly introduces a small increase in Never Valid BT failures for Gemma, perhaps indicating the rootstocks occasionally overwhelm the smaller model's abilities.
\begin{table}[t]
\caption{PyRoboSim failure summary}
\label{tab:pyrobosim_failure_breakdown}
\centering
\footnotesize
\renewcommand{\arraystretch}{1.15}
\setlength{\tabcolsep}{4pt}
\begin{tabular}{llcccc}
\hline
\textbf{Model} & \textbf{Method} & \textbf{Suite} & \textbf{\shortstack{Never Valid\\BT Fail}} & \textbf{\shortstack{Runtime\\Failure}} & \textbf{\shortstack{Goal\\Mismatch}} \\
\hline
Sonnet 4.6  & B1     & Core60 & 0/60 & \textbf{3/60}  & \textbf{1/60}  \\
Sonnet 4.6  & B1     & Lang50 & 0/50 & \textbf{4/50}  & \textbf{1/50}  \\
Sonnet 4.6  & M-Core & Core60 & 0/60 & \textbf{2/60}  & 0/60  \\
Sonnet 4.6  & M-Core & Lang50 & 0/50 & \textbf{2/50}  & 0/50  \\
\hline
Gemma4:31b & B1     & Core60 & \textbf{5/60} & \textbf{22/60} & \textbf{3/60}  \\
Gemma4:31b & B1     & Lang50 & 0/50 & \textbf{15/50} & \textbf{10/50} \\
Gemma4:31b & M-Core & Core60 & \textbf{6/60} & \textbf{3/60}  & \textbf{1/60}  \\
Gemma4:31b & M-Core & Lang50 & \textbf{3/50} & \textbf{5/50}  & \textbf{1/50}  \\
\hline
\end{tabular}
\end{table}

\subsection{Physical Robot Results}\label{sec:experiments:physical_results}
\Cref{tab:hardware_tasks} summarizes the 14 physical robot tasks across four categories of increasing reactive complexity: basic navigation, detection-gated behaviors, track-and-engage, and multi-phase missions with recovery; the two most complex tasks (an indefinite security-guard patrol and a finite greeter mission) serve as extended demonstrations.
\Cref{tab:hardware_failures} reports Valid@1 and Success metrics across these 14 tasks.

\begin{table}[h]
\caption{Panther 14 Task Suite}
\label{tab:hardware_tasks}
\centering
\footnotesize
\renewcommand{\arraystretch}{1.15}
\setlength{\tabcolsep}{4pt}
\begin{tabular}{>{\raggedright\arraybackslash}p{0.18\columnwidth}>{\raggedright\arraybackslash}p{0.72\columnwidth}}
\hline
\textbf{Category} & \textbf{Representative task (paraphrase)} \\
\hline
\multirow{3}{*}{\parbox{0.18\columnwidth}{Navigation\\(n=3)}}
  & \textit{``Go to the kitchen entrance.''} \\
  & \textit{``Go to the vending machine, then to the NRG entrance.''} \\
  & \textit{``Go to the elevator, then return to the NRG entrance.''} \\
\hline
\multirow{4}{*}{\parbox{0.18\columnwidth}{Reactive\\Detection\\(n=4)}}
  & \textit{``Sweep periodically; when a person appears, face them until they leave.''} \\
  & \textit{``Wait until a person is spotted, then navigate to the staircase.''} \\
  & \textit{``Spin slowly until a person is detected, then stop.''} \\
  & \textit{``Patrol between waypoints until a person is spotted; stop and face them.''} \\
\hline
\multirow{3}{*}{\parbox{0.18\columnwidth}{Track \&\\Engage\\(n=3)}}
  & \textit{``When a person appears, turn to face them; resume when they leave.''} \\
  & \textit{``Wait for a person, face them, then drive forward 1 meter.''} \\
  & \textit{``Head to the vending machine; abort if a person is detected, wait, then resume.''} \\
\hline
\multirow{4}{*}{\parbox{0.18\columnwidth}{Multi-phase\\Missions\\(n=4)}}
  & \textit{``Navigate to goal; on failure, back up and retry up to 3 times, then return home.''} \\
  & \textit{``Patrol the hallway; stop and face any detected person, then resume.''} \\
  & \textit{``Patrol continuously between waypoints; face each person encountered.''} \\
  & \textit{``Wait at entrance; pan until a person arrives, greet them, tour waypoints, repeat.''} \\
\hline
\end{tabular}
\end{table}

Regarding validation, Sonnet incurred 1 incorrect XML format validation, where the next retry resulted in a valid BT.
Gemma incurred 1 incorrect XML format validation error and 1 BT.CPP validation error; the subsequent retries resulted in a valid BT. Thus, the developed approach always produced valid BTs for both models considered.

For the physical robot, success is determined subjectively: During experiments, we tested all logical paths of BTs, and observed the robot to perform what the task command stated.
For example, with a task prompt of \textit{``Try to go to the elevator. If navigation fails, back up safely, wait for a bit, and then try again. After you've tried 3 times, come back to the lab entrance.''}, we first prevented the robot from navigating 3 times, observed the recovery behavior, and watched it go back to the entrance. 
Half deductions in \Cref{tab:hardware_failures} are for BTs that slightly altered the intended behavior, e.g. a task prompt stated to \emph{``pan back and forth''}; however, the resulting BT placed a continuous spin node rather than a pan from $-\pi$ to $\pi$, as we would have expected.

For \ref{q:physical_deployment}, the hardware study shows that a contract over opaque ROS2 infrastructure enables the agent to produce valid, executable BTs on a physical robot, with Gemma producing comparable quality to Sonnet despite running locally. 
This convergence at the physical deployment level, combined with the performance gap observed in simulation (Gemma B1: 50\% vs.\ Sonnet B1: 90\% on Core60 Success), illustrates the range of outcomes across the two models and suggests that intermediate models are likely to fall within these bounds.
Interestingly, B1 and M-Core perform comparably on hardware, with B1 slightly outperforming M-Core in some categories.
We interpret this as the hardware rootstocks, designed for a structurally distinct skill schema, provide less compositional leverage than those designed for the simulation domain, suggesting that rootstock benefit is task-dependent.

Regarding the total time to find, call robot MCP server tools, and then construct and send a BT, Sonnet performed this overall process in 50-70 seconds per BT, whereas Gemma running on a local desktop server with 4 NVIDIA RTX 6000 Ada GPUs takes 130-140 seconds per BT. 
In our experimental setup, the client must find and call all MCP tools prior to constructing BTs because each task prompt is sent as a separate Claude Code instance.
For a real-world setting, we envision the client initialized with ``warm robot-server session'' prior to any BT task command.
In preliminary tests, generating a BT after a client has knowledge of robot-server information takes 5-10 seconds which is well-suited for many real-world deployed systems.
 
\begin{table}[h!]
\caption{Panther 14 Valid@1, Success by method \& model.
\checkmark~denotes a perfect score; fractions indicate partial results}
\label{tab:hardware_failures}
\centering
\small
\renewcommand{\arraystretch}{1.3}
\setlength{\tabcolsep}{3pt}
\begin{tabular}{ll cc|cc|cc|cc}
\hline
& & \multicolumn{2}{c}{\textbf{Nav} (3)} & \multicolumn{2}{c}{\textbf{React} (4)} & \multicolumn{2}{c}{\textbf{Track} (3)} & \multicolumn{2}{c}{\textbf{Multi} (4)} \\
\cline{3-10}
\textbf{Method} & \textbf{Model} & V & S & V & S & V & S & V & S \\
\hline
\multirow{2}{*}{M-Core}
& Sonnet & \checkmark & \checkmark & \checkmark & \checkmark & \checkmark & 3/4        & 3/4 & 2.5/4 \\
& Gemma  & 2/3        & \checkmark & \checkmark & \checkmark & \checkmark & \checkmark & 3/4 & 3/4   \\
\hline
\multirow{2}{*}{B1}
& Sonnet & \checkmark & \checkmark & \checkmark & \checkmark & \checkmark & \checkmark & \checkmark & \checkmark \\
& Gemma  & \checkmark & \checkmark & \checkmark & \checkmark & \checkmark & \checkmark & \checkmark & 3.5/4      \\
\hline
\end{tabular}
\end{table}

\subsection{Out-of-Contract Behavior (OOC10)}\label{sec:experiments:ooc_results}
Refusal counts were 7/10 and 5/10 for Sonnet and Gemma on PyRoboSim, and 7/10 for both models on Panther.
In most non-refusal cases, the agent constructed a valid BT for a similar, achievable task.
Non-refusals fall into three failure modes: \emph{approximated} BTs that use the closest available skills (e.g.\ \texttt{Navigate}+\texttt{Wait} for ``call the elevator''), a single \emph{hallucinated} BT (Gemma on PyRoboSim) referencing entities absent from the world vocabulary (\texttt{TV}, \texttt{entrance}, \texttt{lab}), and \emph{chat responses}, observed unanimously on the ``ask the person their name'' prompt, in which the agent broke role and replied conversationally.
The Panther skill set excludes manipulation and exposes only navigation and perception, and we believe this leaves the agent fewer adjacent skills to improvise with than PyRoboSim's broader set, driving higher refusal regardless of LLM.
The in-contract results answer \ref{q:contract_grounding}--\ref{q:physical_deployment}, while OOC10 identifies a refusal behavior limitation discussed further in \Cref{sec:discussion}.

\section{Discussion}\label{sec:discussion}

With this approach, synthesis depends on the quality of the exposed contract, and specifically how the skill schema describes the underlying implementation. 
Additionally, we observe rootstock benefits to be task- and schema-dependent, and they may not transfer.
The OOC10 results show moderate refusal rates, ranging 50--70\% across models and domains indicating the contract alone does not guarantee appropriate refusal for infeasible requests.
The validation gate catches malformed or ungrounded BTs, even if approximated behaviors and conversational responses may require additional intent-level checks.
But with the above validation of the approach we see improving contract quality as an achievable engineering problem potentially including customization for a given domain or task. 

Our architecture is useful in robotics settings where capabilities may be delivered through proprietary navigation stacks, vendor SDKs, or self-contained payload modules.
Compared with agentic systems requiring broad robot source-code access such as RoboClaw~\cite{RoboClaw_2026}, the proposed contract interface reduces the backend source code exposure while still supporting deployable BT synthesis.

Our experiments and results focus on full BT planning for episodic evaluations.
Future work may include a coding agent in the loop during BT execution, enabling online replanning and iterative conversations.
We believe that using a coding agent most effectively for online replanning may require updating world model context -- either in visual, textual, or some other encoded form.

A robot behavior control policy represented as a BT enables layered inspection tiers.
This paper demonstrates a first tier in which formal contract validation is enforced by the robot runtime before execution.
Although not demonstrated, our architecture also enables a next tier in which the agent summarizes the synthesized BT in NL prior to execution, and another tier in which a developer or operator directly inspects the BT structure.
These next tiers are avenues for future work.
\section{Conclusion}\label{sec:conclusion}
This paper presents a contract-grounded coding agent architecture for robot BT synthesis.
By exposing a robot-side contract through an MCP server and enforcing grounding via a validation gate, the architecture enables non-expert operators to synthesize deployable BTs from natural language without knowledge of robot implementation details.
Evaluated across 110 simulated and 14 physical tasks, with both large, API-accessed and small, local models as well as two interface configurations (440 and 56 total BT synthesis trials, respectively), the results show that contract grounding enables near-perfect validity and high task success, that rootstock templates substantially recover success rates on reactive control-flow tasks, and that the approach transfers to a physical ROS2/Nav2 stack.
\section{Appendix}\label{sec:appendix}
Task prompts for the 3 PyRoboSim task suites (Core60, Lang50, OOC10), the Panther 14 task suite, full contracts for both PyRoboSim robot and the Panther robot, and all resulting synthesized BTs are available on the project website: 
\url{https://jsalfity.github.io/agentic-bt-gen-webviewer/}

\printbibliography
\end{document}